\documentclass[10pt,twocolumn,letterpaper]{article}

\usepackage{cvpr}              

\usepackage{graphicx}
\usepackage{amsmath}
\usepackage{amssymb}
\usepackage{booktabs}
\usepackage{placeins}
\usepackage{wrapfig}
\usepackage[export]{adjustbox}
\usepackage{enumitem}
\usepackage{multirow}
\usepackage[accsupp]{axessibility} 
\usepackage{bbding}
\usepackage{pifont}
\usepackage{wasysym}
\usepackage[pagebackref,breaklinks,colorlinks]{hyperref}

\usepackage[capitalize]{cleveref}
\crefname{section}{Sec.}{Secs.}
\Crefname{section}{Section}{Sections}
\Crefname{table}{Table}{Tables}
\crefname{table}{Tab.}{Tabs.}

\def\cvprPaperID{11482} 
\def\confName{CVPR}
\def\confYear{2023}

\begin{document}

\title{Generating Features with Increased Crop-related Diversity \\for Few-Shot Object Detection}

\author{Jingyi Xu\\
Stony Brook University\\
{\tt\small jingyixu@cs.stonybrook.edu}
\and
Hieu Le\\
EPFL\\
{\tt\small minh.le@epfl.ch}
\and
Dimitris Samaras\\
Stony Brook University\\
{\tt\small samaras@cs.stonybrook.edu}
}
\maketitle

\begin{abstract}

Two-stage object detectors generate object proposals and classify them to detect objects in images. 
These proposals often do not  contain the objects perfectly but overlap with them in many possible ways, exhibiting great variability in the difficulty levels of the proposals.
Training a robust classifier against this crop-related variability requires abundant training data, which is not available in few-shot settings. To mitigate this issue, we propose a novel variational autoencoder (VAE) based data generation model, which is capable of generating data with increased crop-related diversity. The main idea is to transform the latent space such latent codes with different norms represent different crop-related variations. This allows us to generate features with increased crop-related diversity in difficulty levels by simply varying the latent norm. In particular, each latent code is rescaled such that its norm linearly correlates with the IoU score of the input crop w.r.t. the ground-truth box. Here the IoU score is a proxy that represents the difficulty level of the crop. We train this VAE model on base classes conditioned on the semantic code of each class and then use the trained model to generate features for novel classes. In our experiments our generated features consistently improve state-of-the-art few-shot object detection methods on the PASCAL VOC and MS COCO datasets.
\end{abstract}

\section{Introduction}
\label{sec:intro}
\begin{figure}[!th]
\begin{center}
\includegraphics[width=0.97\linewidth]{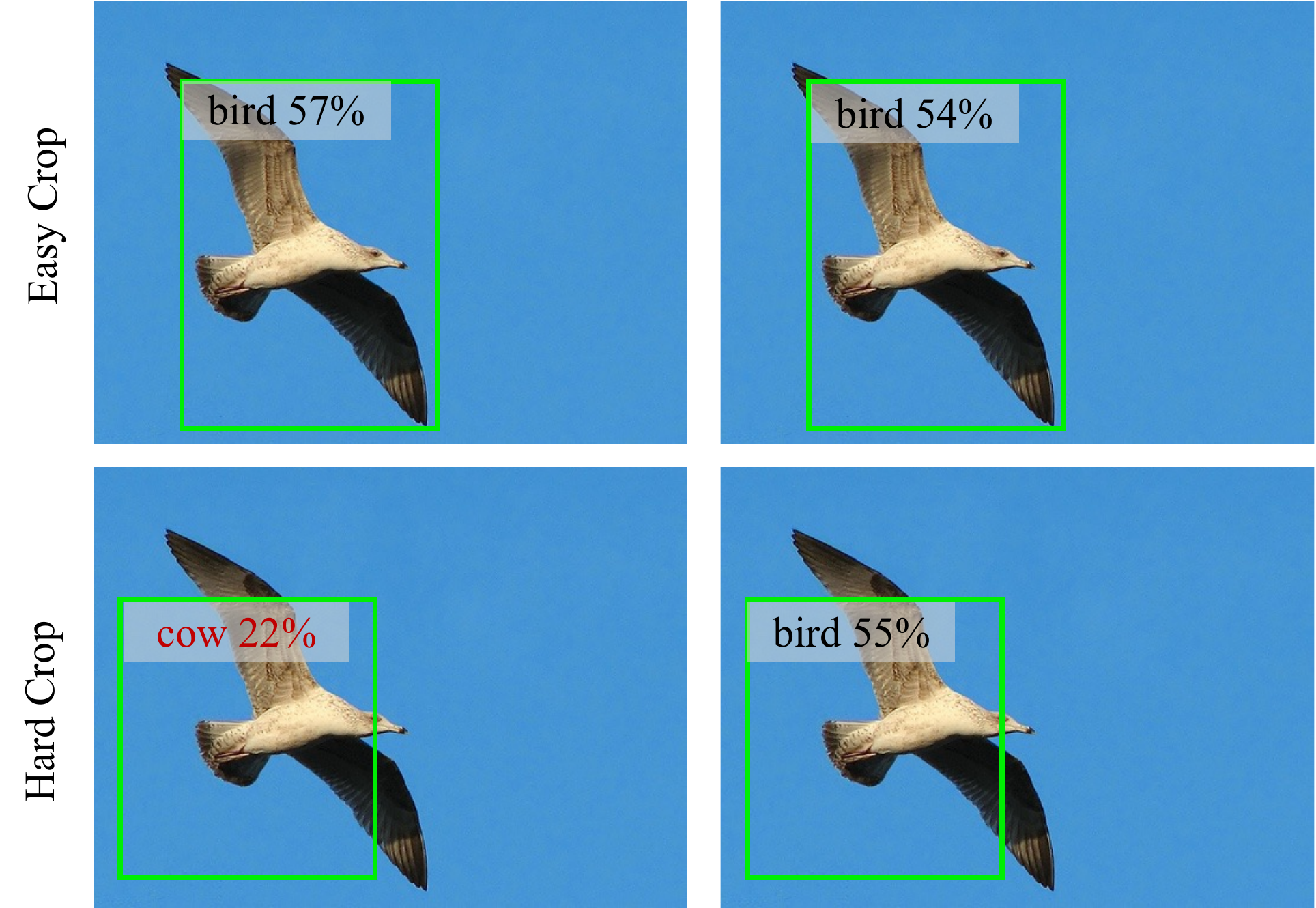}
\makebox[0.05\linewidth]{ }
\makebox[0.43\linewidth]{(a) DeFRCN \cite{Qiao2021DeFRCNDF}  }
\makebox[0.43\linewidth]{(b) Ours  }
\caption{\textbf{Robustness to different object crops of the same object instance}. (a) The classifier head of the state-of-the-art FSOD method \cite{Qiao2021DeFRCNDF} classifies correctly a simple crop of the bird but misclassifies a hard crop where some parts are missing. (b) Our method can handle this case since it is trained with additional generated features with increased crop-related diversity. We show the class with the highest confidence score.} \vspace{-5mm}
\label{fig:teaser}
\end{center}
\end{figure}

Object detection plays a vital role in many computer vision systems. However, training a robust object detector often requires a large amount of training data with accurate bounding box annotations. 
Thus, there has been increasing attention on few-shot object detection (FSOD), which learns to detect novel object categories from just a few annotated training samples. It is particularly useful for problems where annotated data can be hard and costly to obtain such as rare medical conditions~\cite{Ouyang2020SelfSupervisionWS, Wang2021FewShotLB}, rare animal species \cite{Le_RS_penguin_22,cub}, satellite images \cite{Borowicz2019AerialtrainedDL,Le_2019_CVPR_Workshops}, or failure cases in autonomous driving systems \cite{Rezaei2020ZeroshotLA,Majee2021FewShotLF,Majee2021MetaGM}. 


For the most part, state-of-the-art FSOD methods are built on top of a two-stage framework \cite{faster-rcnn}, which includes a region proposal network that generates multiple image crops from the input image and a classifier that labels these proposals. While the region proposal network generalizes well to novel classes, the classifier is more error-prone due to the lack of training data diversity \cite{Sun2021FSCEFO}. To mitigate this issue, a natural approach is to generate additional features for novel classes \cite{Zhang2021HallucinationIF,Zhu2020DontEL,Hayat2020SynthesizingTU}. For example, Zhang \etal~\cite{Zhang2021HallucinationIF} 
propose a feature hallucination network to use the variation from base classes to diversify training data for novel classes. 
For zero-shot detection (ZSD), Zhu \etal~\cite{Zhu2020DontEL} propose to synthesize visual features for unseen objects based on a conditional variational auto-encoder. 
%
%
Although much progress has been made, the lack of data diversity is still a challenging issue for FSOD methods.

Here we discuss a specific type of data diversity that greatly affects the accuracy of FSOD algorithms. 
Specifically, given a test image, the classifier needs to accurately classify multiple object proposals\footnote{Note that an RPN typically outputs 1000 object proposals per image.} that overlap the object instance in various ways. The features of these image crops exhibit great variability induced by different object scales, object parts included in the crops, object positions within the crops, and backgrounds. We observe a typical scenario where the state-of-the-art FSOD method, DeFRCN \cite{Qiao2021DeFRCNDF}, only classifies correctly a few among many proposals overlapping an object instance of a few-shot class. In fact, different ways of cropping an object can result in features with various difficulty levels. An example is shown in Figure \ref{fig:teaser}a where the image crop shown in the top row is classified correctly while another crop shown in the bottom row confuses the classifier due to some missing object parts.  In general, the performance of the method on those hard cases is significantly worse than on easy cases (see section \ref{sec:hardcases}). However, building a classifier robust against crop-related variation is challenging since there are only a few images per few-shot class.

In this paper, we propose a novel data generation method to mitigate this issue.
Our goal is to generate features with diverse crop-related variations for the few-shot classes and use them as additional training data to train the classifier. 
Specifically, we aim to obtain a diverse set of features whose difficulty levels vary from easy to hard \textit{w.r.t.} how the object is cropped.\footnote{In this paper, the difficulty level is strictly related to how the object is cropped.}  
To achieve this goal, we design our generative model such that it allows us to control the difficulty levels of the generated samples. Given a model that generates features from a latent space, our main idea is to enforce that the magnitude of the latent code linearly correlates with the difficulty level of the generated feature, \textit{i.e.}, the latent code of a harder feature is placed further away from the origin and vice versa. In this way, we can control the difficulty level by simply changing the norm of the corresponding latent code.   

In particular, our data generation model is based on a conditional variational autoencoder (VAE) architecture. The VAE consists of an encoder that maps the input to a latent representation and a decoder that reconstructs the input from this latent code. In our case, inputs to the VAE are  object proposal features, extracted from a pre-trained object detector. The goal is to associate the norm (magnitude) of the latent code with the difficulty level of the object proposal. To do so, we rescale the latent code such that its norm linearly correlates with the Intersection-Over-Union (IoU) score of the input object proposal \textit{w.r.t.} the ground-truth object box. This IoU score is a proxy that partially indicates the difficulty level: A high IoU score indicates that the object proposal significantly overlaps with the object instance while a low IoU score indicates a harder case where a part of the object can be missing. With this rescaling step, we can bias the decoder to generate harder samples by increasing the latent code magnitude and vice versa. In this paper, we use latent codes with different norms varying from small to large to obtain a diverse set of features which can then serve as additional training data for the few-shot classifier. 



To apply our model to FSOD, we first train our VAE model using abundant data from the base classes. The VAE is conditioned on the semantic code of the input instance category. After the VAE model is trained, we use the semantic embedding of the few-shot class as the conditional code to synthesize new features for the corresponding class. 
In our experiments, we use our generated samples to fine-tune the baseline few-shot object detector - DeFRCN \cite{Qiao2021DeFRCNDF}.
Surprisingly, a vanilla conditional VAE model trained with only  ground-truth box features brings a $3.7\%$ nAP50 improvement over the  DeFRCN baseline in the 1-shot setting of the PASCAL VOC dataset \cite{Everingham2009ThePV}. Note that we are the first FSOD method using VAE-generated features to support the training of the classifier.  Our proposed Norm-VAE can further improve this new state-of-the-art by another $2.1\%$, \textit{i.e.}, from $60\%$ to $62.1\%$. In general, the generated features from Norm-VAE consistently improve the state-of-the-art few-shot object detector \cite{Qiao2021DeFRCNDF} for both PASCAL VOC and MS COCO \cite{Lin2014MicrosoftCC} datasets.

Our main contributions can be summarized as follows: 
\begin{itemize}
\setlength\itemsep{-.3em}
    \item We show that lack of crop-related diversity in training data of novel classes is a crucial problem for FSOD.
    \item We propose Norm-VAE, a novel VAE architecture that can effectively increase crop-related diversity in difficulty levels into the generated samples to support the training of FSOD classifiers. 
    \item Our experiments show that the object detectors trained with our additional features achieve state-of-the-art FSOD in both PASCAL VOC and MS COCO datasets.
\end{itemize}

\section{Related Work}


\label{sec:rw}

\textbf{Few-shot Object Detection} Few-shot object detection aims to detect novel classes from limited annotated examples of previously unseen classes. A number of prior methods 
\cite{Kaul22,Han_2021_ICCV,Perez-Rua_2020_CVPR,Fan_2020_CVPR,Zhu2021SemanticRR,Sun2021FSCEFO,Li_2021_CVPR,Fan_2021_CVPR, Wu2021GeneralizedAD,Wu2021UniversalPrototypeEF,Wu2022InstanceInvariantDA,Kaul22,han2022few,Ma2022FewShotEO,Fan2022FewShotMC,Ma2022MutuallyRS,LeICCV2017}
have been proposed to address this challenging task. One line of work focuses on the \textbf{meta-learning} paradigm, which has been widely explored in few-shot classification \cite{Kang2019FewShotOD,Yan2019MetaRT,Xiao2020FewShotOD,Schwartz2019RepMetRM,Yang2020RestoringNI,Yang2020ContextTransformerTO,Fan2020FewShotOD,Wang2019MetaLearningTD}. Meta-learning based approaches introduce a meta-learner to acquire meta-knowledge that can be then transferred to novel classes. \cite{Kang2019FewShotOD} propose a meta feature learner and a reweighting module to fully exploit generalizable features from base classes and quickly adapt the prediction network to predict novel classes. \cite{Wang2019MetaLearningTD} propose specialized meta-strategies to disentangle the learning of category-agnostic and category-specific components in a CNN based detection model. Another line of work adopts a \textbf{two-stage fine-tuning} strategy and has shown great potential recently \cite{Wang2020FrustratinglySF,Wu2020MultiScalePS,Sun2021FSCEFO,Cao2021FewShotOD,Qiao2021DeFRCNDF}. \cite{Wang2020FrustratinglySF} propose to fine-tune only box classifier and box regressor with novel data while freezing the other paramters of the model. This simple stragetegy outperforms previous meta-learners. FSCE \cite{Sun2021FSCEFO} leverages a contrastive proposal encoding loss to promote instance level intra-class compactness and inter-class variance. Orthogonal to existing work, we propose to generate new samples for FSOD. Another\textbf{ data generation based} method for FSOD is Halluc \cite{Zhang2021HallucinationIF}. 
However, their method learns to transfer the shared within-class variation from base classes while we focus on the crop-related variance.

\textbf{Feature Generation} Feature generation has been widely used in low-shot learning tasks. The common goal is to generate reliable and diverse additional data. For example, in image classification, \cite{Xu2022FS} propose to generate representative samples using a VAE model conditioned on the semantic embedding of each class. The generated samples are then used together with the original samples to construct class prototypes for few-shot learning. In spirit, their conditional-VAE system is similar to ours. \cite{Xian2019FVAEGAND2AF} propose to combine a VAE and a Generative Adversarial Network (GAN) by sharing the decoder of VAE and generator of GAN to synthesize features for zero-shot learning. In the context of object detection, \cite{Zhang2021HallucinationIF} propose to transfer the shared modes of within-class variation from base classes
to novel classes to hallucinate new samples. \cite{Zhu2021SemanticRR} propose to synthesize visual features for unseen objects from semantic information and
augment existing training algorithms to incorporate unseen object detection. Recently, \cite{Huang2022RobustRF} propose to synthesize samples which are both intra-class diverse and inter-class separable to support the training of zero-shot object detector.
However, these methods do not take into consideration the variation induced by different crops of the same object, which is the main focus of our proposed method.

\textbf{Variational Autoencoder} Different VAE variants have been proposed to generate diverse data \cite{beta_vae, Klys2018LearningLS,Via2021PROPERTYCV,Shao2020ControlVAECV}. $\beta$-VAE \cite{beta_vae} imposes a heavy penalty on the KL divergence term to enhance the disentanglement of the latent dimensions. By traversing the values of latent variables, $\beta$-VAE can generate data with disentangled variations. 
ControlVAE \cite{Shao2020ControlVAECV} improves upon $\beta$-VAE by introducing a controller to automatically tune the hyperparameter added in the VAE objective.
However, disentangled representation learning can not capture the desired properties without supervision. 
Some VAE methods allow explicitly controllable feature generation including CSVAE \cite{Klys2018LearningLS} and PCVAE \cite{Via2021PROPERTYCV}. CSVAE \cite{Klys2018LearningLS} 
learns latent dimensions associated with binary properties. The learned latent subspace can easily be inspected and independently manipulated. PCVAE \cite{Via2021PROPERTYCV} uses a Bayesian model to inductively bias the latent representation. Thus, moving along the learned latent dimensions can control specific properties of the generated data. Both CSVAE and PCVAE use additional latent variables and enforce additional constrains to control properties.
In contrast, our Norm-VAE directly encodes a variational factor into the norm of the latent code. Experiments show that our strategy outperforms other VAE architectures, while being simpler and without any additional training components.

\section{Method}
\label{sec:method}
In this section, we first review the problem setting of few-shot object detection and the conventional two-stage fine-tuning framework. Then we introduce our method that tackles few-shot object detection via generating features with increased crop-related diversity.

\subsection{Preliminaries}
In few-shot object detection, the training set is divided into a base set $D^B$ with abundant annotated instances of classes $C^B$, and a novel set $D^N$ with few-shot data of classes $C^N$, where $C^B$ and $C^N$ are non-overlapping. For a sample $(x, y) \in D^B \cup D^N$, $x$ is the input image and $y = \{(c_i, b_i), i = 1,...,n\}$ denotes the categories $c \in C^B \cup C^N $ and bounding box coordinates $b$ of the $n$ object instances in the image $x$. The number of objects for each class in $C^N$ is $K$ for $K$-shot detection. We aim to obtain a few-shot detection model with the ability to detect objects in the test set with classes in $C^B \cup C^N$.

Recently, two-stage fine-tuning methods have shown great potential in improving few-shot detection. In these two-stage detection frameworks, a Region Proposal Network (RPN) takes the output feature maps from a backbone feature extractor as inputs and generates region proposals. A Region-of-Interest (RoI) head feature extractor first pools the region proposals to a fixed size and then encodes them as vector embeddings, known as the RoI features. A classifier is trained on top of the RoI features to classify the categories of the region proposals.

The fine-tuning often follows a simple two-stage training pipeline, \textit{i.e.}, the data-abundant base training stage and the novel fine-tuning stage. In the base training stage, the model collects transferable knowledge across a large base set with sufficient annotated data. Then in the fine-tuning stage, it performs quick adaptation on the novel classes with limited data. 
Our method aims to generate features with diverse crop-related variations to enrich the training data for the classifier head during the fine-tuning stage.
In our experiments, we show that our generated features significantly improve the performance of DeFRCN \cite{Qiao2021DeFRCNDF}. 

\subsection{Overall Pipeline}
\definecolor{chromeyellow}{rgb}{1, 0.65, 0}
\definecolor{myblue}{rgb}{0.15, 0.32, 0.83}
\begin{figure*}[t]
\begin{center}
\includegraphics[width=0.95\linewidth]{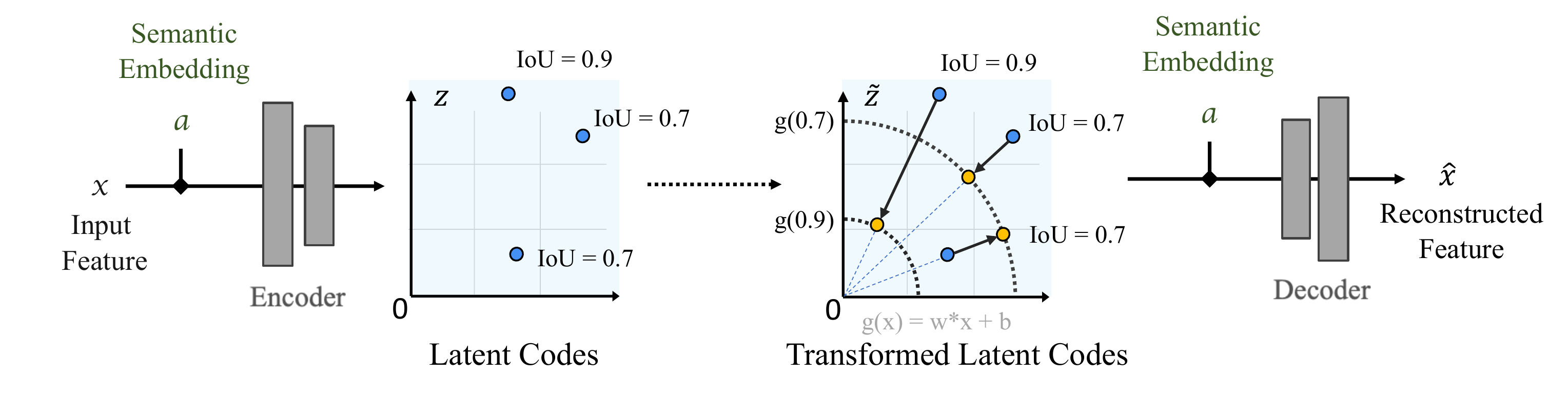} \vspace{-5mm}
\end{center}
\caption{\textbf{Norm-VAE for modelling crop-related variations.} 
The original latent code $z$ is rescaled to $\hat{z}$ such that the norm of $\hat{z}$ linearly correlates with the IoU score of the input crop (\textit{w.r.t.} the ground truth box). The original latent codes are colored in {\color{myblue}\textbf{blue}} while the rescaled ones are colored in {\color{chromeyellow}\textbf{yellow}}. The norm of the new latent code is the output of a simple linear function $g(\cdot)$ taking the IoU score as the single input. As can be seen, the two points whose IoU = 0.7 are both rescaled to norm $g(0.7)$ while another point whose IoU = 0.9 is mapped to norm $g(0.9)$. As a result, different latent norms represent different crop-related variations, enabling diverse feature generation. 
} \vspace{-1.5mm}
\label{fig:overview}
\end{figure*}

Figure \ref{fig:overview} summarizes the main idea of our proposed VAE model. For each input object crop, we first use a pre-trained object detector to obtain its RoI feature. The encoder takes as input the RoI feature and the semantic embedding of the input class to output a latent code $z$. We then transform $z$ such that its norm linearly correlates with the IoU score of the input object crop \textit{w.r.t.} the ground-truth box. The new norm is the output of a simple linear function $g(\cdot)$ taking the IoU score as the single input. The decoder takes as input the new latent code and the class semantic embedding to output the reconstructed feature. Once the VAE is trained, we use the semantic embedding of the few-shot class as the conditional code to synthesize new features for the class. To ensure the diversity \textit{w.r.t.} object crop in generated samples, we vary the norm of the latent code when generating features. The generated features are then used together with the few-shot samples to fine-tune the object detector.


\subsubsection{Norm-VAE for Feature Generation}
 We develop our feature generator based on a conditional VAE architecture \cite{CVAE}. Given an input object crop, we first obtain its Region-of-Interest (RoI)  feature $f$ via a pre-trained object detector. The RoI feature $f$ is the input for the VAE. The VAE is composed of an Encoder $E(f, a)$, which maps a visual feature $f$ to a latent code $z$, and a decoder $G(z, a)$ which reconstructs the feature $f$ from $z$. Both $E$ and $G$ are conditioned on the class semantic embedding $a$. We obtain this class semantic embedding $a$ by inputting the class name into a semantic model \cite{Miller1992WordNetAL,Radford2021LearningTV_CLIP}. It contains class-specific information and serves as a controller to determine the categories of the generated samples. Conditioning on these semantic embeddings allows reliably generating features for the novel classes based on the learned information from the base classes \cite{Xu2022FS}. Here we assume that the class names of both base and novel classes are available and we can obtain the semantic embedding of all classes. 
 
We first start from a vanilla conditional VAE model. The loss function for training this VAE for a feature $f_i$ of class $j$ can be defined as:
 \begin{equation}\label{eq:cvae}
  \begin{aligned}
      L_{V}(f_i) =  \textnormal{KL} \left( q(z_i|f_i,a^j)||p(z|a^j) \right)  - \\ \textnormal{E}_{q(z_i|f_i, a^j)}[\textnormal{log }p(f_i|z_i,a^j)],
  \end{aligned}
\end{equation}
where $a^j$ is the semantic embedding of class $j$. The first term is the Kullback-Leibler divergence between the VAE posterior $q(z|f,a)$ and a prior distribution $p(z|a)$. The second term is the decoder's reconstruction error. $q(z|f,a)$ is modeled as $E(f, a)$ and $p(f|z,a)$ is equal to $G(z, a)$. The prior distribution is assumed to be $\mathcal{N}(0,I)$ for all classes. 


The goal is to control the crop-related variation in a generated sample. Thus, we establish a direct correspondence between the latent norm and the crop-related variation.
To accomplish this, we transform the latent code such that its norm correlates with the IoU score of the input crop. Given an input RoI feature $f_i$ of a region with an IoU score $s_i$, we first input this RoI feature to the encoder to obtain its latent code $z_i$. We then transform $z_i$ to $\tilde{z_i}$ such that the norm of $\tilde{z_i}$  correlates to $s_i$. The new latent code $\tilde{z_i}$  is the output of the transformation function $\mathcal{T}(\cdot,\cdot)$:  
\begin{equation}\label{eq:normalization}
    \tilde{z_i}  = \mathcal{T}(z_i,s_i) = \frac{z_i} {\lVert z_i \rVert} * {g}(s_i),
\end{equation}
where $\lVert z_i \rVert$ is the $L_2$ norm of $z_i$, $s_i$ is the IoU score of the input proposal \textit{w.r.t.} its ground-truth object box, and ${g}(\cdot)$ is a simple pre-defined linear function that maps an IoU score to a norm value. 
With this new transformation step, the loss function of the VAE from equation \ref{eq:cvae} for an input feature $f_i$ from class $j$ with an IoU score $s_i$ thus can be rewritten as:
\\
 \begin{equation}\label{eq:norm_vae}
 \begin{aligned}
      L_{V}(f_i,s_i) =  \textnormal{KL} \left( q(z_i|f_i,a^j)||p(z|a^j) \right)  - \\ \textnormal{E}_{q\left({z_i}|f_i, a^j\right)}\left[\textnormal{log }p(f_i|\mathcal{T}(z_i,s_i),a^j)\right].
 \end{aligned}
\end{equation}


\subsubsection{Generating Diverse Data for Improving Few-shot Object Detection}
After the VAE is trained on the base set, we generate a set of features with the trained decoder. Given a class $y$ with a semantic vector $a^y$ and a noise vector $z$, we generate a set of augmented features $\mathbb{G}^y$:
 \begin{equation}\label{eq:output}
 \begin{aligned}
    \mathbb{G}^y = \{\hat{f}|\hat{f} = G(\frac{z} {\lVert z \rVert} * \beta, a^y)\},
 \end{aligned}
\end{equation}
where we vary $\beta$ to obtain generated features with more crop-related variations. The value range of $\beta$ is chosen based on the mapping function $g(\cdot)$. The augmented features are used together with the few-shot samples to fine-tune the object detector. We fine-tune the whole system using an additional classification loss computed on the generated features together with the original losses computed on real images. This is much simpler than the previous method of \cite{Zhang2021HallucinationIF} where they fine-tune their system via an EM-like (expectation-maximization) manner.

\section{Experiments}
\label{sec:experiments}

\subsection{Datasets and Evaluation Protocols}

We conduct experiments on both PASCAL VOC (07 + 12) \cite{Everingham2009ThePV} and MS COCO datasets \cite{Lin2014MicrosoftCC}. For fair comparison, we follow the data split construction and evaluation protocol used in previous works \cite{Kang2019FewShotOD}. The PASCAL VOC dataset contains 20 categories. We use the same 3 base/novel splits with TFA \cite{Wang2020FrustratinglySF} and refer them as Novel Split 1,2, 3. Each split contains 15 base classes and 5 novel classes. Each novel class has $K$ annotated instances, where $K = 1,2,3,5,10$. We report AP50 of the novel categories (nAP50) on VOC07 test set. For MS COCO, the 60 categories disjoint with PASCAL VOC are used as base classes while the remaining 20 classes are used as novel classes. We evaluate our method on shot 1,2,3,5,10,30 and COCO-style AP of the novel classes is adopted as the evaluation metrics.

\subsection{Implementation Details}
\label{sec:implement}
Feature generation methods like ours in theory can be built on top of many few-shot object detectors. In our experiments, we use the pre-trained Faster-RCNN \cite{faster-rcnn} with ResNet-101 \cite{resnet} following previous work DeFRCN \cite{Qiao2021DeFRCNDF}. The dimension of the extracted RoI feature is $2048$. For our feature generation model, the encoder consists of three fully-connected (FC) layers and the decoder consists of two FC layers, both with $4096$ hidden units. LeakyReLU and ReLU are the non-linear activation functions in the hidden and output layers, respectively. The dimensions of the latent space and the semantic vector are both set to be $512$. 
Our semantic embeddings are extracted from a pre-trained CLIP \cite{Radford2021LearningTV_CLIP} model in all main experiments. An additional experiment using Word2Vec \cite{Mikolov2013EfficientEO} embeddings is reported in Section \ref{sec:word2vec}. After the VAE is trained on the base set with various augmented object boxes , we use the trained decoder to generate $k=30$ features per class and incorporate them into the fine-tuning stage of the DeFRCN model. We set the function $g(\cdot)$ in Equation \ref{eq:normalization} to a simple linear function $g(x) = w *x + b$ which maps an input IoU score $x$ 
to the norm of the new latent code. Note that $x$ is in range $[0.5,1]$ and the norm of the latent code of our VAE before the rescaling typically centers around $\sqrt{512}$ ($512$ is the dimension of the latent code). We empirically choose $g(\cdot)$ such that the new norm ranges from $\sqrt{512}$ to $5 * \sqrt{512}$. 
We provide further analyses on the choice of $g(\cdot)$ in the supplementary material.
For each feature generation iteration, we gradually increase the value of the controlling parameter $\beta$ in Equation \ref{eq:output} 
with an interval of $0.75$.


\subsection{Few-shot Detection Results}
\label{sec:FS_results}

  \begin{table*}[!h] 
  \centering
\resizebox{0.96\textwidth}{!}{%
  \begin{tabular}{l|ccccc|ccccc|ccccc}
    \toprule
     & \multicolumn{5}{c|}{Novel Split 1} & \multicolumn{5}{c|}{Novel Split 2} &
              \multicolumn{5}{c}{Novel Split 3}  \\ Method
    & 1& 2& 3& 5& 10& 1& 2& 3& 5& 10& 1& 2& 3& 5& 10 \\
    \midrule 
    TFA w/ fc \cite{Wang2020FrustratinglySF} & 36.8 & 29.1 & 43.6 & 55.7 & 57.0 & 18.2  & 29.0 & 33.4 & 35.5 & 39.0 & 27.7 & 33.6 & 42.5 & 48.7 & 50.2 \\
    TFA w/ cos \cite{Wang2020FrustratinglySF} & 39.8 & 36.1 & 44.7 & 55.7 & 56.0 & 23.5  & 26.9 & 34.1 & 35.1 & 39.1 & 30.8 & 34.8 & 42.8 & 49.5 & 49.8 \\
    MPSR \cite{Wu2020MultiScalePS} & 41.7 & - & 51.4 & 55.2 & 61.8 & 24.4  & - & 39.2 & 35.1 & 39.9 & 47.8 & - & 42.3 & 48.0 & 49.7 \\
    FsDetView \cite{Xiao2020FewShotOD} & 24.2 & 35.3 & 42.2 & 49.1 & 57.4 & 21.6  & 24.6 & 31.9 & 37.0 & 45.7 & 21.2 & 30.0 & 37.2 & 43.8 & 49.6 \\
    FSCE \cite{Sun2021FSCEFO} & 44.2 & 43.8 & 51.4 & 61.9 & 63.4 & 27.3  & 29.5 & 43.5 & 44.2 & 50.2 & 37.2 & 41.9 & 47.5 & 54.6 & 58.5 \\
    CME \cite{Li2021BeyondMC} & 41.5 & 47.5 & 50.4 & 58.2 & 60.9 & 27.2  & 30.2 & 41.4 & 42.5 & 46.8 & 34.3 & 39.6 & 45.1 & 48.3 & 51.5 \\
    SRR-FSD \cite{Zhu2021SemanticRR} & 47.8 & 50.5 & 51.3 & 55.2 & 56.8 & 32.5 & 35.3 & 39.1 & 40.8 & 43.8 & 40.1 & 41.5 & 44.3 & 46.9 & 46.4 \\
    Halluc. \cite{Zhang2021HallucinationIF} & 45.1 & 44.0 & 44.7 & 55.0 & 55.9 & 23.2 & 27.5 & 35.1 & 34.9 & 39.0 & 30.5 & 35.1 & 41.4 & 49.0 & 49.3 \\
    FSOD-MC \cite{Fan2022FewShotMC} & 40.1 & 44.2 & 51.2 & 62.0 & 63.0 & 33.3 & 33.1 & 42.3 & 46.3 & 52.3 & 36.1 & 43.1 & 43.5 & 52.0 & 56.0 \\
    FADI \cite{Cao2021FewShotOD} & 50.3 & 54.8 & 54.2 & 59.3 & 63.2 & 30.6  & 35.0 & 40.3 & 42.8 & 48.0 & 45.7 & 49.7 & 49.1 & 48.3 & 51.5 \\
    CoCo-RCNN \cite{Ma2022FewShotEO} & 43.9 & 44.5 & 53.1 & 64.6 & 65.5 & 29.4 & 31.3 & 43.8 & 44.3 & 51.8 & 39.1 & 43.9 & 47.2 & 54.7 & 60.3 \\
    MRSN \cite{Ma2022MutuallyRS} & 47.6 & 48.6 & 57.8 & 61.9 & 62.6 & 31.2 & 38.3 & 46.7 & 47.1 & 50.6 & 35.5 & 30.9 & 45.6 & 54.4 & 57.4 \\
    FCT \cite{han2022few} & 49.9 & 57.1 & 57.9 & 63.2 & 67.1 & 27.6 & 34.5 & 43.7 & 49.2 & 51.2 & 39.5 & 54.7 & 52.3 & 57.0 & 58.7 \\
    Pseudo-Labelling \cite{Kaul22}& 54.5 & 53.2 & 58.8 & 63.2 & 65.7 & 32.8  & 29.2 & 50.7 & 49.8 & 50.6 & 48.4 & 52.7 & 55.0 & 59.6 & 59.6 \\
    DeFRCN \cite{Qiao2021DeFRCNDF}  & 56.3 & 60.3 & 62.0 & 67.0 & 66.1 & 35.7   & 45.2 & 51.5 & 54.1 & 53.3 & 54.5 & 55.6 & 56.6 & 60.8 & 62.7 \\ \midrule
    Vanila-VAE (Ours) & 60.0 & 63.3 & 66.3 & 68.3 & 67.1 & 39.3 & 46.2 & 52.7 & 53.5 & 53.4 & 56.0 & 58.8 & 57.1 & 62.6 & 63.6 \\
    Norm-VAE (Ours)  & \textbf{62.1} & \textbf{64.9} & \textbf{67.8} & \textbf{69.2} & \textbf{67.5}  & \textbf{39.9} & \textbf{46.8} &  \textbf{54.4} & \textbf{54.2} & \textbf{53.6} & \textbf{58.2} & \textbf{60.3} & \textbf{61.0} & \textbf{64.0} & \textbf{65.5} \\
    \bottomrule
  \end{tabular}}  \\ \vspace{6pt}
  \caption{\textbf{Few-shot object detection performance (nAP50) on PASCAL VOC dataset}. We evaluate the performance on three different splits. Our method consistently improves upon the baseline for all three splits across all shots.
  Best performance  in bold.
  }\label{tab:voc}%
  \vspace{2pt}
 \end{table*}

\begin{table*}[!h] 
  \centering
\resizebox{0.7\textwidth}{!}{%
  \begin{tabular}{l|cccccc|cccccc}
    \toprule
     & \multicolumn{6}{c|}{nAP}  &
              \multicolumn{6}{c}{nAP75}  \\
    Method & 1& 2& 3& 5& 10& 30& 1& 2& 3& 5& 10& 30 \\
    \midrule 
    TFA w/ fc \cite{Wang2020FrustratinglySF} & 2.9 & 4.3 & 6.7 & 8.4 & 10.0 & 13.4 &  2.8 & 4.1 & 6.6 & 8.4 & 9.2 & 13.2 \\
    TFA w/ cos \cite{Wang2020FrustratinglySF} & 3.4 & 4.6 & 6.6 & 8.3 & 10.0 & 13.7 & 3.8 & 4.8 & 6.5 & 8.0 & 9.3 & 13.2 \\
    MPSR \cite{Wu2020MultiScalePS} & 2.3 & 3.5 & 5.2 & 6.7 & 9.8 & 14.1 & 2.3 & 3.4 & 5.1 & 6.4 & 9.7 & 14.2 \\
    FADI \cite{Cao2021FewShotOD} & 5.7 & 7.0 & 8.6 & 10.1 & 12.2 & 16.1 &  6.0 & 7.0 & 8.3 & 9.7 & 11.9 & 15.8 \\
    FCT \cite{han2022few} & - & 7.9 & - & - & 17.1 & 21.4 & - & 7.9 & - & - & 17.0 & 22.1 \\
    Pseudo-Labelling \cite{Kaul22} \dag & - & - & - & - & 17.8 & \textbf{24.5} & - & - & - & - & \textbf{17.8} & \textbf{25.0} \\
    DeFRCN \cite{Qiao2021DeFRCNDF}  & 6.6 & 11.7 & 13.3 & 15.6 & 18.7 & 22.4 & 7.0 & 12.2 & 13.6 & 15.1 & 17.6 & 22.2 \\ \midrule
    Vanilla-VAE (ours) & 8.8 & 13.0 & 14.1 & \textbf{15.9} & \textbf{18.7} & {22.5} & 7.9 & 12.5 & 13.4 & 15.1 & 17.6 & 22.2 \\
    Norm-VAE (ours)   & \textbf{9.5} & \textbf{13.7} & \textbf{14.3} & \textbf{15.9} & \textbf{18.7}  & 
    {22.5} & \textbf{8.8} & \textbf{13.7} & \textbf{14.2} & \textbf{15.3} & \textbf{17.8} & {22.4} \\
    \bottomrule
  \end{tabular}} \\ \vspace{2.5pt}
  \caption{\textbf{Few-shot detection performance for the novel classes on MS COCO dataset}. Our approach outperforms baseline methods in most cases, especially in low-shot settings ($K<10$). $\dag $ applies mosaic data augmentation introduced in \cite{Bochkovskiy2020YOLOv4OS} during fine-tuning. 
  Best performance  in bold.
  }\label{tab:coco}%
\end{table*}

We use the generated features from our VAE model together with the few-shot samples to fine-tune DeFRCN. We report the performance of two models: ``Vanilla-VAE'' denotes the performance of the model trained with generated features from a vanilla VAE trained on the base set of ground-truth bounding boxes and ``Norm-VAE'' denotes the performance of the model trained with features generated from our proposed Norm-VAE model.

\textbf{PASCAL VOC} Table \ref{tab:voc} shows our results for all three random novel splits from PASCAL VOC. Simply using a VAE model trained with the original data outperforms the state-of-the-art method DeFRCN in all shot and split on PASCAL VOC benchmark. In particular, vanilla-VAE improves DeFRCN by $3.7\%$ for 1-shot and $4.3\%$ for 3-shot on Novel Split 1. Using additional data from our proposed Norm-VAE model consistently improves the results across all settings. We provide qualitative examples in the supplementary material.

\textbf{MS COCO} Table \ref{tab:coco} shows the FSOD results on MS COCO dataset. Our generated features bring significant improvements in most cases, especially in low-shot settings (K $\leq$ 10). For example, Norm-VAE brings a $2.9\%$ and a $2.0\%$ nAP improvement over DeFRCN in 1-shot and 2-shot settings, respectively. Pseudo-Labeling is better than our method in higher shot settings. However, they apply mosaic data augmentation \cite{Bochkovskiy2020YOLOv4OS} during fine-tuning. 

\section{Analyses}
\label{sec:ana}

\subsection{Effectiveness of Norm-VAE}
\label{sec:effectiveness}
We compare the performance of Norm-VAE with a baseline vanilla VAE model that is trained with the same set of augmented data. 
As shown in Table \ref{tab:norm_vae}, using the vanilla VAE with more training data does not bring performance improvement compared to the VAE model trained with the base set. This suggests that training with more diverse data does not guarantee diversity in generated samples \textit{w.r.t.} a specific property. Our method, by contrast, improves the baseline model by $1.3\% \sim 1.9\%$, which demonstrates the effectiveness of our proposed Norm-VAE. 
\begin{table*}[!h] 
  \centering
\resizebox{0.8\textwidth}{!}{%
  \begin{tabular}{l|c|ccc|ccc|ccc}
    \toprule
    \multirow{2}{*} {Method} & {Semantic } & \multicolumn{3}{c|}{Novel Split 1} & \multicolumn{3}{c|}{Novel Split 2} & \multicolumn{3}{c}{Novel Split 3}\\
    
    & Embedding & 1-shot & 2-shot & 3-shot & 1-shot & 2-shot & 3-shot & 1-shot & 2-shot & 3-shot \\
    \midrule 
    DeFRCN \cite{Qiao2021DeFRCNDF} & - & {56.3} & {60.3} & {62.0} & {35.7} & {45.2} & {51.5} & {54.5} & {55.6} & {56.6} \\
    \midrule 
    Vanilla VAE & \multirow{2}{*}{Word2Vec} & {60.4} & {62.9} & \textbf{66.7} & {38.7} & {45.2} & {52.9} & {55.6} & {58.7} & {57.9} \\
    Norm-VAE & & \textbf{61.6} & \textbf{63.4} & {66.3} & \textbf{40.7} & \textbf{46.4} & \textbf{53.3} & \textbf{56.8} & \textbf{59.0} & \textbf{60.2} \\
    \midrule 
    Vanilla VAE & \multirow{2}{*}{CLIP} & {60.0} & {63.3} & {66.3} & {39.3} & {46.2} & {52.7} & {56.0} & {58.8} & {57.1} \\
    Norm-VAE & & \textbf{62.1} & \textbf{64.9} & \textbf{67.8} & \textbf{39.9} & \textbf{46.8} & \textbf{54.4} & \textbf{58.2} & \textbf{60.3} & \textbf{61.0} \\
    \bottomrule
  \end{tabular}}
  \\ \vspace{6pt}
  \caption{\textbf{FSOD Performance of VAE models trained with different class semantic embeddings}. CLIP \cite{Radford2021LearningTV_CLIP} is trained with 400M pairs (image and its text title) collected from the web while Word2Vec \cite{Mikolov2013EfficientEO} is trained with only text data.
  }\label{tab:word2vec}%
\end{table*}

\begin{table}[!h] 
  \centering
\resizebox{0.42\textwidth}{!}{%
  \begin{tabular}{l|c|c|c|c}
    \toprule
     & {Data} & {1-shot} & {2-shot} & {3-shot} \\
    \midrule
    DeFRCN \cite{Qiao2021DeFRCNDF} & - & 56.3 & 60.3  & 62.0 \\
    VAE & Orginal & 60.0 & 63.3  & {66.3} \\
    VAE & Augmented & 60.1 & 62.7  & {66.4} \\
    \midrule
    Norm-VAE & Augmented & \textbf{62.1} & \textbf{64.9} & \textbf{67.8} \\
    \bottomrule
  \end{tabular}} \\ \vspace{6pt}
  \caption{\textbf{Performance comparisons between vanilla VAE and Norm-VAE on PASCAL VOC dataset}. Training a the vanilla VAE with the augmented data does not bring performance improvement. One possible reason is that the generated samples are not guaranteed to be diverse even with sufficient data.
  }\label{tab:norm_vae}%
\end{table}

\subsection{Performance Using Different Semantic Embeddings}
\label{sec:word2vec}
We use CLIP \cite{Radford2021LearningTV_CLIP} features in our main experiments. In Table \ref{tab:word2vec}, we compare this model with another model trained with Word2Vec \cite{Mikolov2013EfficientEO} on PASCAL VOC dataset. Note that CLIP model is trained with 400M pairs (image and its text title) collected from the web while Word2Vec is trained with only text data. Our Norm-VAE trained with Word2Vec embedding achieves similar performance to the model trained with CLIP embedding. In both cases, the model outperform the state-of-the-art FSOD method in all settings.

\subsection{Robustness against Inaccurate Localization}
 \label{sec:robustness}

In this section, we conduct experiments to show that our object detector trained with features with diverse crop-related variation is more robust against  inaccurate bounding box localization. Specifically, we randomly select 1000 testing instances from PASCAL VOC test set and create 30 augmented boxes for each ground-truth box. Each augmented box is created by enlarging the ground-truth boxes by $x\%$ for each dimension where $x$ ranges from 0 to 30. The result is summarized in Figure \ref{fig:confi_iou} where ``Baseline'' denotes the performance of DeFRCN\cite{Qiao2021DeFRCNDF}, ``VAE'' is the performance of the model trained with features generated from a vanilla VAE, and ``Norm-VAE'' is the model trained with generated features from our proposed model.  

Figure \ref{fig:confi_iou} (a) shows the classification accuracy of the object detector on the augmented box as the IoU score between the augmented bounding box and the ground-truth box decreases. For both the baseline method DeFRCN  and the model trained with features from a vanilla VAE, the accuracy drops by $\sim 10\%$ as the IoU score decreases from 1.0 to 0.5. These results suggest that these models perform much better for boxes that have higher IoU score \textit{w.r.t.} the ground-truth boxes. 
Our proposed method has higher robustness to these inaccurate boxes: the accuracy of the model trained with features from Norm-VAE 
only drops by $\sim 5\%$ when IoU score decreases from 1 to 0.5. 

Figure \ref{fig:confi_iou} (b) plots the average probability score of the classifier on the ground-truth category as the IoU score decreases. Similarly, the probability score of both baseline DeFRCN and the model trained with features from a vanilla VAE drops around 0.08 as the IoU score decreases from 1.0 to 0.5. The model trained with features from Norm-VAE, in comparison, has more stable probability score as the IoU threshold decreases.

  \def\subboxsize{\textwidth}
 \begin{figure}[ht!]

\resizebox{0.48\textwidth}{!}{
 \centering
\hspace{-0.3cm}\includegraphics[width=\linewidth]{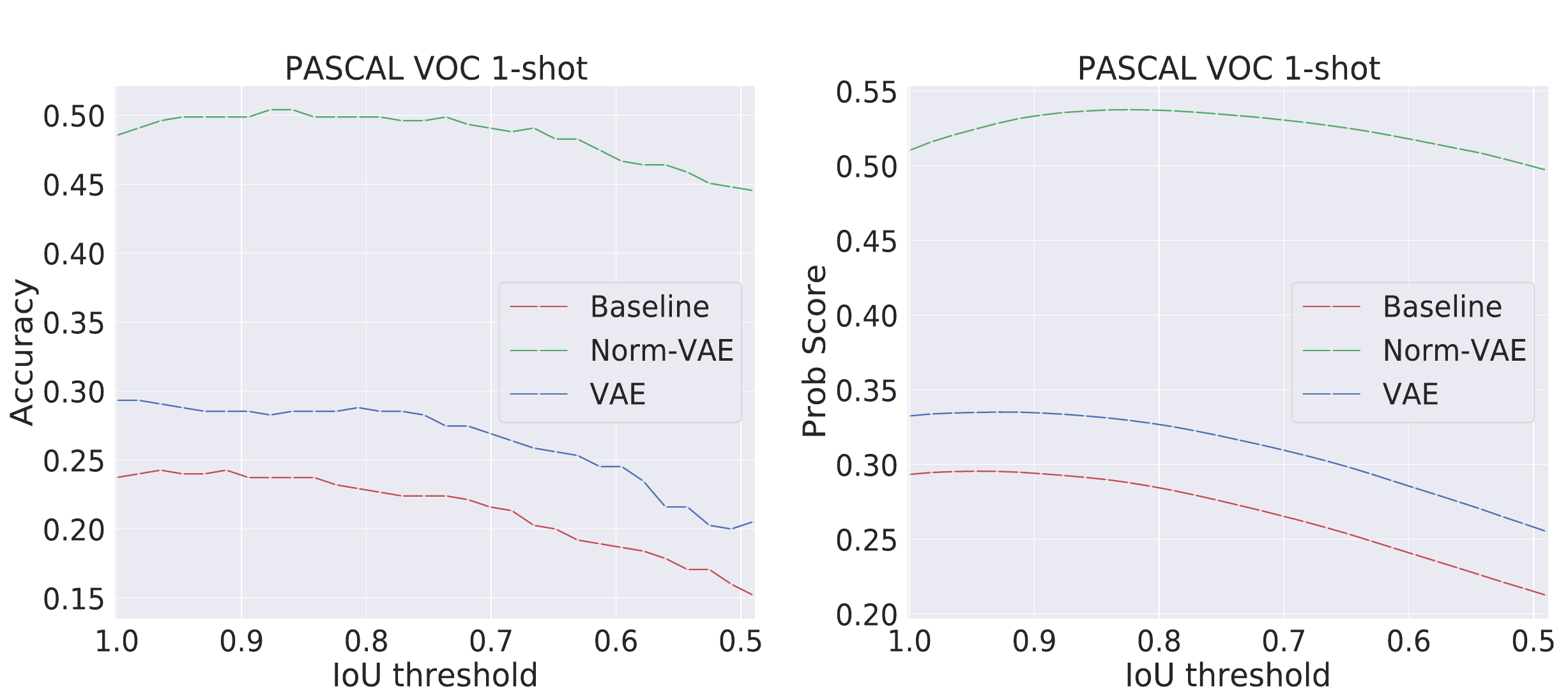}}
\makebox[0.23\subboxsize]{\hspace{0.6cm}(a) Accuracy}
\makebox[0.23\subboxsize]{\hspace{0.8cm} (b) Probability score}
 \caption{ \textbf{Classification accuracy and probability score of the object detector on the augmented box}. We compare between the baseline DeFRCN \cite{Qiao2021DeFRCNDF}, the model trained with features from vanilla VAE and our proposed Norm-VAE. By generating features with diverse crop-related variations, we increase the object detector's robustness against inaccurate object box localization.
}
    \label{fig:confi_iou}
\end{figure}

\begin{table}[!h] 
  \centering
\resizebox{0.48\textwidth}{!}{%
  \begin{tabular}{l|c|c|c}
    \toprule
    Method & {1-shot} & {2-shot} & {3-shot} \\
    \midrule 
    DeFRCN\cite{Qiao2021DeFRCNDF} & {16.6} & {13.3}  & {15.2} \\
    Ours ($\uparrow$ Improvement) & {18.8 ($\uparrow$2.2)}  & {16.4 ($\uparrow3.1$)} & {19.2 ($\uparrow$4.0)}  \\
    \bottomrule
  \end{tabular}} \\ \vspace{10pt}
  \caption{\textbf{AP50$\sim$75 of our method and DeFRCN on PASCAL VOC dataset}. AP 50$\sim$75 refers to the average precision computed on the proposals with the IoU thresholds between $50\%$ and $75\%$ and discard the proposals with IoU scores larger than 0.75, i.e., only ``\textit{hard}'' cases. 
  }\label{tab:ap50_75}%
\end{table}

\subsection{Performance on Hard Cases}
\label{sec:hardcases}
In Table \ref{tab:ap50_75}, we show AP 50$\sim$75 of our method on PASCAL VOC dataset (Novel Split 1) in comparison with the state-of-the-art method DeFRCN. Here AP 50$\sim$75 refers to the average precision computed on the proposals with the IoU thresholds between $50\%$ and $75\%$ and discard the proposals with IoU scores (\textit{w.r.t.} the ground-truth box) larger than 0.75. Thus, AP 50$\sim$75 implies the performance of the model in ``\textit{hard}'' cases where the proposals do not significantly overlap the ground-truth object boxes. 
In this extreme test, the performance of both models are worse than their AP50 counterparts (Table \ref{tab:voc}), showing that FSOD methods are generally not robust to those hard cases. Our method mitigates this issue, outperforming DeFRCN by substantial margins. However, the performance is still far from perfect. Addressing these challenging cases is a fruitful venue for future FSOD work. 

\begin{table*}[!h] 
  \centering
\resizebox{0.82\textwidth}{!}{%
  \begin{tabular}{l|cc|cc|cc|cc}
    \toprule
    Features & \multicolumn{2}{c|}{1-shot} & \multicolumn{2}{c|}{2-shot} & \multicolumn{2}{c|}{3-shot} & \multicolumn{2}{c}{5-shot}\\
    & nAP50 & nAP75 & nAP50 & nAP75 & nAP50 & nAP75 & nAP50 & nAP75\\
    \midrule
    {Low-IoU (Hard cases)} & \textbf{60.9} & \textcolor{black}{30.5} & \textbf{63.7} & \textcolor{black}{40.6} & \textbf{66.6} & \textcolor{black}{40.7} & \textbf{68.9} & \textcolor{black}{41.2}  \\
    {High-IoU (Easy cases)} & \textcolor{black}{60.2} & \textbf{31.6} & \textcolor{black}{63.2} & \textbf{41.0} & \textcolor{black}{66.3} & \textbf{41.5} & \textcolor{black}{68.3} & \textbf{42.1}  \\
    \bottomrule
  \end{tabular}}  \\ \vspace{6pt}
  \caption{ \textbf{Comparison between models trained with different groups of generated features}. The model trained with ``Low-IoU'' (hard cases) features has better nAP50 scores while the ``High-IoU'' (easy cases) model has better nAP75 scores.
  Features corresponding to different difficulty levels improve the performance differently in terms of nAP50 and nAP75.
  }\label{tab:iou}%
  \vspace{-10pt}
\end{table*}

\subsection{Performance with Different Subsets of Generated Features}
In this section, we conduct experiments to show that different groups of generated features affect the performance of the object detector differently.
Similar to Section \ref{sec:implement}, we generate 30 new features per few-shot class with various latent norms. However, instead of using all norms, we only use large norms (top 30\% highest values) to generate the first group of features and only small norms (top 30\% lowest values) to generate the second group of features. During training, larger norms correlate to input crops with smaller IoU scores \textit{w.r.t.} the ground-truth boxes and vice versa. Thus, we denote these two groups as ``Low-IoU'' and ``High-IoU'' correspondingly. We train two models using these two sets of features and compare their performance in Table \ref{tab:iou}. As can be seen, the model trained with ``Low-IoU'' features has higher AP50 while the ``High-IoU'' model has higher AP75 score. This suggests that different groups of features affect the performance of the classifier differently. The ``Low-IoU'' features tend to increase the model's robustness to hard-cases  while the ``High-IoU'' features can improve the performance for easier cases. Note that the performance of both of these models is not as good as the model trained with diverse variations and interestingly, very similar to the performance of the vanilla VAE model (Table \ref{tab:voc}). 

\subsection{Comparisons with Other VAE architectures}
\label{sec:other_VAE}

Our proposed Norm-VAE can increase diversity \textit{w.r.t.} image crops in generated samples.  
Here, we compare the performance of our proposed Norm-VAE with other VAE architectures, including $\beta$-VAE \cite{beta_vae} and CSVAE~\cite{Klys2018LearningLS}. 
We train all models on image features of augmented object crops on PASCAL VOC dataset using the same backbone feature extractor. 
For $\beta$-VAE, we generate additional features by traversing a randomly selected dimension of the latent code. For CSVAE, we manipulate the learned latent subspace to  enforce variations in the generated samples. We use generated features from each method to fine-tune DeFRCN. The results are summarized in Table \ref{tab:vae_variants}. In all cases, the generated features greatly benefit the baseline DeFRCN. This shows that lacking crop-related variation is a critical issue for FSOD, and augmenting features with increased crop-related diversity can effectively alleviate the problem.  %
Our proposed Norm-VAE outperforms both $\beta$-VAE and CSVAE in all settings. Note that CSVAE requires additional encoders to learn a pre-defined subspace correlated with the property, while our Norm-VAE directly encode this into the latent norm without any additional constraints.

\begin{table}[!h] 
  \centering
\resizebox{0.32\textwidth}{!}{%
  \begin{tabular}{l|c|c|c}
    \toprule
      & {1-shot} & {2-shot} & {3-shot} \\
    \midrule
    DeFRCN \cite{Qiao2021DeFRCNDF} & 56.3 & 60.3  & 62.0 \\
    \midrule
    $\beta$-VAE\cite{beta_vae} &  61.3 & 64.0  & {67.3} \\
    CSVAE\cite{Klys2018LearningLS}  & 61.6 & 64.1  & {67.4} \\
    Norm-VAE  & \textbf{62.1} & \textbf{64.9} & \textbf{67.8} \\
    \bottomrule
  \end{tabular}}  \\ \vspace{2pt}
  \caption{\textbf{ Comparison between Norm-VAE and other VAE variants.} Norm-VAE outperforms $\beta$-VAE and CSVAE on PASCAL VOC dataset under all settings. 
  Best performance in bold.
  }\label{tab:vae_variants}%
  \vspace{-14pt}
\end{table}

\section{Conclusion and Future Works}
We tackle the lack of crop-related variability in the training data of FSOD, which makes the model not robust to different object proposals of the same object instance.
To this end, we propose a novel VAE model that can generate features with increased crop-related diversity.  Experiments show that such increased diversity in the generated samples significantly improves the current state-of-the-art FSOD performance for both PASCAL VOC and MS COCO datasets. Our proposed VAE model is simple, easy to implement, and allows modifying the difficulty levels of the generated samples. In general, generative models whose outputs can be manipulated according to different properties, are crucial to various frameworks and applications. 
In future work, we plan to address the following limitations of our work:
1) We bias the decoder to increase the diversity in generated samples instead of explicitly enforcing it. 2) Our proposed method is designed to generate visual features of object boxes for FSOD. Generating images  might be required in other applications. Another direction to extend our work is to represent other variational factors in the embedding space to effectively diversify generated data.

\newcommand{\myheading}[1]{\vspace{1ex}\noindent \textbf{#1}}
\myheading{Acknowledgements.} 
This research was partially supported by NSF grants IIS-2123920 and IIS-2212046 and the NASA Biodiversity program (Award 80NSSC21K1027).

\newpage

\FloatBarrier
{\small
\bibliographystyle{ieee_fullname}
\bibliography{egbib}

\begin{thebibliography}{10}\itemsep=-1pt

\bibitem{Bochkovskiy2020YOLOv4OS}
Alexey Bochkovskiy, Chien-Yao Wang, and Hong-Yuan~Mark Liao.
\newblock Yolov4: Optimal speed and accuracy of object detection.
\newblock {\em ArXiv}, abs/2004.10934, 2020.

\bibitem{Borowicz2019AerialtrainedDL}
Alex Borowicz, Hieu Le, Grant Humphries, G. Nehls, Caroline H{\"o}schle, V.
  Kosarev, and H. Lynch.
\newblock Aerial-trained deep learning networks for surveying cetaceans from
  satellite imagery.
\newblock {\em PLoS ONE}, 14, 2019.

\bibitem{Cao2021FewShotOD}
Yuhang Cao, Jiaqi Wang, Ying Jin, Tong Wu, Kai Chen, Ziwei Liu, and Dahua Lin.
\newblock Few-shot object detection via association and discrimination.
\newblock In {\em NeurIPS}, 2021.

\bibitem{Everingham2009ThePV}
Mark Everingham, Luc~Van Gool, Christopher K.~I. Williams, John~M. Winn, and
  Andrew Zisserman.
\newblock The pascal visual object classes (voc) challenge.
\newblock {\em International Journal of Computer Vision}, 88, 2009.

\bibitem{Fan2022FewShotMC}
Qi Fan, Chi-Keung Tang, and Yu-Wing Tai.
\newblock Few-shot object detection with model calibration.
\newblock In {\em ECCV}, 2022.

\bibitem{Fan2020FewShotOD}
Qi Fan, Wei Zhuo, and Yu-Wing Tai.
\newblock Few-shot object detection with attention-rpn and multi-relation
  detector.
\newblock In {\em CVPR}, pages 4012--4021, 2020.

\bibitem{Fan_2020_CVPR}
Qi Fan, Wei Zhuo, Chi-Keung Tang, and Yu-Wing Tai.
\newblock Few-shot object detection with attention-rpn and multi-relation
  detector.
\newblock In {\em CVPR}, June 2020.

\bibitem{Fan_2021_CVPR}
Zhibo Fan, Yuchen Ma, Zeming Li, and Jian Sun.
\newblock Generalized few-shot object detection without forgetting.
\newblock In {\em CVPR}, June 2021.

\bibitem{Via2021PROPERTYCV}
Xiaojie Guo, Yuanqi Du, and Liang Zhao.
\newblock Property controllable variational autoencoder via and invertible
  mutual dependence.
\newblock In {\em ICLR}, 2021.

\bibitem{Han_2021_ICCV}
Guangxing Han, Yicheng He, Shiyuan Huang, Jiawei Ma, and Shih-Fu Chang.
\newblock Query adaptive few-shot object detection with heterogeneous graph
  convolutional networks.
\newblock In {\em ICCV}, October 2021.

\bibitem{han2022few}
Guangxing Han, Jiawei Ma, Shiyuan Huang, Long Chen, and Shih-Fu Chang.
\newblock Few-shot object detection with fully cross-transformer.
\newblock In {\em CVPR}, 2022.

\bibitem{Hayat2020SynthesizingTU}
Nasir Hayat, Munawar Hayat, Shafin Rahman, Salman~Hameed Khan, Syed~Waqas
  Zamir, and Fahad~Shahbaz Khan.
\newblock Synthesizing the unseen for zero-shot object detection.
\newblock In {\em ACCV}, 2020.

\bibitem{resnet}
Kaiming He, Xiangyu Zhang, Shaoqing Ren, and Jian Sun.
\newblock Deep residual learning for image recognition.
\newblock In {\em CVPR}, June 2016.

\bibitem{beta_vae}
Irina Higgins, Loic Matthey, Arka Pal, Christopher Burgess, Xavier Glorot,
  Matthew Botvinick, Shakir Mohamed, and Alexander Lerchner.
\newblock beta-vae: Learning basic visual concepts with a constrained
  variational framework.
\newblock In {\em ICLR}, 2017.

\bibitem{Huang2022RobustRF}
Peiliang Huang, Junwei Han, De Cheng, and Dingwen Zhang.
\newblock Robust region feature synthesizer for zero-shot object detection.
\newblock In {\em CVPR}, 2022.

\bibitem{Kang2019FewShotOD}
Bingyi Kang, Zhuang Liu, Xin Wang, Fisher Yu, Jiashi Feng, and Trevor Darrell.
\newblock Few-shot object detection via feature reweighting.
\newblock {\em ICCV}, 2019.

\bibitem{Kaul22}
Prannay Kaul, Weidi Xie, and Andrew Zisserman.
\newblock Label, verify, correct: A simple few-shot object detection method.
\newblock In {\em CVPR}, 2022.

\bibitem{Klys2018LearningLS}
Jack Klys, Jake Snell, and Richard~S. Zemel.
\newblock Learning latent subspaces in variational autoencoders.
\newblock In {\em NeurIPS}, 2018.

\bibitem{Le_2019_CVPR_Workshops}
Hieu Le, Bento Goncalves, Dimitris Samaras, and Heather Lynch.
\newblock Weakly labeling the antarctic: The penguin colony case.
\newblock In {\em CVPR Workshops}, June 2019.

\bibitem{Le_RS_penguin_22}
Hieu Le, Dimitris Samaras, and Heather~J. Lynch.
\newblock A convolutional neural network architecture designed for the
  automated survey of seabird colonies, 2022.

\bibitem{LeICCV2017}
Hieu Le, Chen-Ping Yu, Gregory Zelinsky, and Dimitris Samaras.
\newblock Co-localization with category-consistent features and geodesic
  distance propagation.
\newblock In {\em ICCV Workshop}, 2017.

\bibitem{Li2021BeyondMC}
Bohao Li, Boyu Yang, Chang Liu, Feng Liu, Rongrong Ji, and Qixiang Ye.
\newblock Beyond max-margin: Class margin equilibrium for few-shot object
  detection.
\newblock 2021.

\bibitem{Li_2021_CVPR}
Yiting Li, Haiyue Zhu, Yu Cheng, Wenxin Wang, Chek~Sing Teo, Cheng Xiang,
  Prahlad Vadakkepat, and Tong~Heng Lee.
\newblock Few-shot object detection via classification refinement and
  distractor retreatment.
\newblock In {\em CVPR}, June 2021.

\bibitem{Lin2014MicrosoftCC}
Tsung-Yi Lin, Michael Maire, Serge~J. Belongie, James Hays, Pietro Perona, Deva
  Ramanan, Piotr Doll{\'a}r, and C.~Lawrence Zitnick.
\newblock Microsoft coco: Common objects in context.
\newblock In {\em ECCV}, 2014.

\bibitem{Ma2022FewShotEO}
Jiawei Ma, Guangxing Han, Shiyuan Huang, Yuncong Yang, and Shih-Fu Chang.
\newblock Few-shot end-to-end object detection via constantly concentrated
  encoding across heads.
\newblock In {\em ECCV}, 2022.

\bibitem{Ma2022MutuallyRS}
TianXue Ma, Mingwei Bi, Jian Zhang, Wang Yuan, Zhizhong Zhang, Yuan Xie,
  Shouhong Ding, and Lizhuang Ma.
\newblock Mutually reinforcing structure with proposal contrastive consistency
  for few-shot object detection.
\newblock In {\em ECCV}, 2022.

\bibitem{Majee2021FewShotLF}
Anay Majee, Kshitij Agrawal, and A. Subramanian.
\newblock Few-shot learning for road object detection.
\newblock {\em ArXiv}, abs/2101.12543, 2021.

\bibitem{Majee2021MetaGM}
Anay Majee, A. Subramanian, and Kshitij Agrawal.
\newblock Meta guided metric learner for overcoming class confusion in few-shot
  road object detection.
\newblock {\em ArXiv}, abs/2110.15074, 2021.

\bibitem{Mikolov2013EfficientEO}
Tomas Mikolov, Kai Chen, Gregory~S. Corrado, and Jeffrey Dean.
\newblock Efficient estimation of word representations in vector space.
\newblock In {\em ICLR}, 2013.

\bibitem{Miller1992WordNetAL}
George~A. Miller.
\newblock Wordnet: A lexical database for english.
\newblock {\em Commun. ACM}, 38:39--41, 1992.

\bibitem{Ouyang2020SelfSupervisionWS}
Cheng Ouyang, Carlo Biffi, Chen Chen, Turkay Kart, Huaqi Qiu, and Daniel
  Rueckert.
\newblock Self-supervision with superpixels: Training few-shot medical image
  segmentation without annotation.
\newblock In {\em ECCV}, 2020.

\bibitem{Perez-Rua_2020_CVPR}
Juan-Manuel Perez-Rua, Xiatian Zhu, Timothy~M. Hospedales, and Tao Xiang.
\newblock Incremental few-shot object detection.
\newblock In {\em CVPR}, June 2020.

\bibitem{Qiao2021DeFRCNDF}
Limeng Qiao, Yuxuan Zhao, Zhiyuan Li, Xi Qiu, Jianan Wu, and Chi Zhang.
\newblock Defrcn: Decoupled faster r-cnn for few-shot object detection.
\newblock In {\em ICCV}, 2021.

\bibitem{Radford2021LearningTV_CLIP}
Alec Radford, Jong~Wook Kim, Chris Hallacy, Aditya Ramesh, Gabriel Goh,
  Sandhini Agarwal, Girish Sastry, Amanda Askell, Pamela Mishkin, Jack Clark,
  Gretchen Krueger, and Ilya Sutskever.
\newblock Learning transferable visual models from natural language
  supervision.
\newblock In {\em ICML}, 2021.

\bibitem{faster-rcnn}
Shaoqing Ren, Kaiming He, Ross~B. Girshick, and Jian Sun.
\newblock Faster r-cnn: Towards real-time object detection with region proposal
  networks.
\newblock {\em IEEE Transactions on Pattern Analysis and Machine Intelligence},
  39, 2015.

\bibitem{Rezaei2020ZeroshotLA}
Mahdi Rezaei and Mahsa Shahidi.
\newblock Zero-shot learning and its applications from autonomous vehicles to
  covid-19 diagnosis: A review.
\newblock {\em Intelligence-Based Medicine}, 3:100005 -- 100005, 2020.

\bibitem{Schwartz2019RepMetRM}
Eli Schwartz, Leonid Karlinsky, Joseph Shtok, Sivan Harary, Mattias Marder,
  Sharath Pankanti, Rog{\'e}rio~Schmidt Feris, Abhishek Kumar, Raja Giryes, and
  Alexander~M. Bronstein.
\newblock Repmet: Representative-based metric learning for classification and
  few-shot object detection.
\newblock In {\em CVPR}, 2019.

\bibitem{Shao2020ControlVAECV}
Huajie Shao, Shuochao Yao, Dachun Sun, Aston Zhang, Shengzhong Liu, Dongxin
  Liu, Jun Wang, and Tarek~F. Abdelzaher.
\newblock Controlvae: Controllable variational autoencoder.
\newblock In {\em ICML}, 2020.

\bibitem{CVAE}
Kihyuk Sohn, Honglak Lee, and Xinchen Yan.
\newblock Learning structured output representation using deep conditional
  generative models.
\newblock In {\em NIPS}, 2015.

\bibitem{Sun2021FSCEFO}
Bo Sun, Banghuai Li, Shengcai Cai, Ye Yuan, and Chi Zhang.
\newblock Fsce: Few-shot object detection via contrastive proposal encoding.
\newblock In {\em CVPR}, 2021.

\bibitem{Wang2021FewShotLB}
Wenji Wang, Qing Xia, Zhiqiang Hu, Zhennan Yan, Zhuowei Li, Yang Wu, Ning
  Huang, Yue Gao, Dimitris~N. Metaxas, and Shaoting Zhang.
\newblock Few-shot learning by a cascaded framework with shape-constrained
  pseudo label assessment for whole heart segmentation.
\newblock {\em IEEE Transactions on Medical Imaging}, 40:2629--2641, 2021.

\bibitem{Wang2020FrustratinglySF}
Xin Wang, Thomas~E. Huang, Trevor Darrell, Joseph Gonzalez, and Fisher Yu.
\newblock Frustratingly simple few-shot object detection.
\newblock {\em ArXiv}, abs/2003.06957, 2020.

\bibitem{Wang2019MetaLearningTD}
Yu-Xiong Wang, Deva Ramanan, and Martial Hebert.
\newblock Meta-learning to detect rare objects.
\newblock In {\em ICCV}, 2019.

\bibitem{cub}
P. Welinder, S. Branson, T. Mita, C. Wah, F. Schroff, S. Belongie, and P.
  Perona.
\newblock {Caltech-UCSD Birds 200}.
\newblock Technical Report CNS-TR-2010-001, California Institute of Technology,
  2010.

\bibitem{Wu2021UniversalPrototypeEF}
Aming Wu, Yahong Han, Linchao Zhu, and Yi Yang.
\newblock Universal-prototype enhancing for few-shot object detection.
\newblock {\em ICCV}, pages 9547--9556, 2021.

\bibitem{Wu2022InstanceInvariantDA}
Aming Wu, Yahong Han, Linchao Zhu, and Yi Yang.
\newblock Instance-invariant domain adaptive object detection via progressive
  disentanglement.
\newblock {\em IEEE Transactions on Pattern Analysis and Machine Intelligence},
  44:4178--4193, 2022.

\bibitem{Wu2021GeneralizedAD}
Aming Wu, Suqi Zhao, Cheng Deng, and Wei Liu.
\newblock Generalized and discriminative few-shot object detection via
  svd-dictionary enhancement.
\newblock In {\em NeurIPS}, 2021.

\bibitem{Wu2020MultiScalePS}
Jiaxi Wu, Songtao Liu, Di Huang, and Yunhong Wang.
\newblock Multi-scale positive sample refinement for few-shot object detection.
\newblock {\em ArXiv}, abs/2007.09384, 2020.

\bibitem{Xian2019FVAEGAND2AF}
Yongqin Xian, Saurabh Sharma, Bernt Schiele, and Zeynep Akata.
\newblock F-vaegan-d2: A feature generating framework for any-shot learning.
\newblock In {\em CVPR}, 2019.

\bibitem{Xiao2020FewShotOD}
Yang Xiao and Renaud Marlet.
\newblock Few-shot object detection and viewpoint estimation for objects in the
  wild.
\newblock In {\em ECCV}, 2020.

\bibitem{Xu2022FS}
Jingyi Xu and Hieu Le.
\newblock Generating representative samples for few-shot classification.
\newblock In {\em CVPR}, 2022.

\bibitem{Yan2019MetaRT}
Xiaopeng Yan, Ziliang Chen, Anni Xu, Xiaoxi Wang, Xiaodan Liang, and Liang Lin.
\newblock Meta r-cnn: Towards general solver for instance-level low-shot
  learning.
\newblock In {\em ICCV}, 2019.

\bibitem{Yang2020RestoringNI}
Yukuan Yang, Fangyun Wei, Miaojing Shi, and Guoqi Li.
\newblock Restoring negative information in few-shot object detection.
\newblock {\em ArXiv}, abs/2010.11714, 2020.

\bibitem{Yang2020ContextTransformerTO}
Ze Yang, Yali Wang, Xianyu Chen, Jianzhuang Liu, and Yu Qiao.
\newblock Context-transformer: Tackling object confusion for few-shot
  detection.
\newblock In {\em AAAI}, 2020.

\bibitem{Zhang2021HallucinationIF}
Weilin Zhang and Yu-Xiong Wang.
\newblock Hallucination improves few-shot object detection.
\newblock {\em CVPR}, pages 13003--13012, 2021.

\bibitem{Zhu2021SemanticRR}
Chenchen Zhu, Fangyi Chen, Uzair Ahmed, Zhiqiang Shen, and Marios Savvides.
\newblock Semantic relation reasoning for shot-stable few-shot object
  detection.
\newblock In {\em CVPR}, 2021.

\bibitem{Zhu2020DontEL}
Pengkai Zhu, Hanxiao Wang, and Venkatesh Saligrama.
\newblock Don’t even look once: Synthesizing features for zero-shot
  detection.
\newblock In {\em CVPR}, 2020.

\end{thebibliography}


\begin{thebibliography}{1}\itemsep=-1pt

\bibitem{Meng2021MagFaceAU}
Qiang Meng, Shichao Zhao, Zhida Huang, and Feng Zhou.
\newblock Magface: A universal representation for face recognition and quality
  assessment.
\newblock {\em 2021 IEEE/CVF Conference on Computer Vision and Pattern
  Recognition (CVPR)}, pages 14220--14229, 2021.

\bibitem{Qiao2021DeFRCNDF}
Limeng Qiao, Yuxuan Zhao, Zhiyuan Li, Xi Qiu, Jianan Wu, and Chi Zhang.
\newblock Defrcn: Decoupled faster r-cnn for few-shot object detection.
\newblock In {\em ICCV}, 2021.

\end{thebibliography}
}

\end{document}


\title{Generating Features with Increased Crop-related Diversity \\for Few-Shot Object Detection\\
Supplementary Material}

\author{Jingyi Xu\\
Stony Brook University\\
{\tt\small jingyixu@cs.stonybrook.edu}
\and
Hieu Le\\
EPFL\\
{\tt\small minh.le@epfl.ch}
\and
Dimitris Samaras\\
Stony Brook University\\
{\tt\small samaras@cs.stonybrook.edu}
}
\maketitle

\section{Overview}
In this document, we provide additional experiments and analyses. In particular:
\begin{itemize}
    \item Section \ref{sec:visualization_inaccurate_box} provides visualizations of the detection results for inaccurate bounding boxes.
    \item Section \ref{sec:number_of_generated_samples} provides the results of using different numbers of additional samples to fine-tine the model.
    \item Section \ref{sec:appendix_vis_2} provides additional visualizations of the detection results. 
    \item Section \ref{sec:mapping_function} shows the impact of the mapping function on the final results.
    \item Section \ref{sec:augmented_data} provides the details of how we generate additional training data.
\end{itemize}

 \section{Detection Results for Inaccurate Bounding Boxes}
 \label{sec:visualization_inaccurate_box}
 \def\subboxsize{0.3\textwidth}
 \begin{figure*}[!ht]
 \centering
\hspace*{-0.1cm}\includegraphics[width=0.8\linewidth]{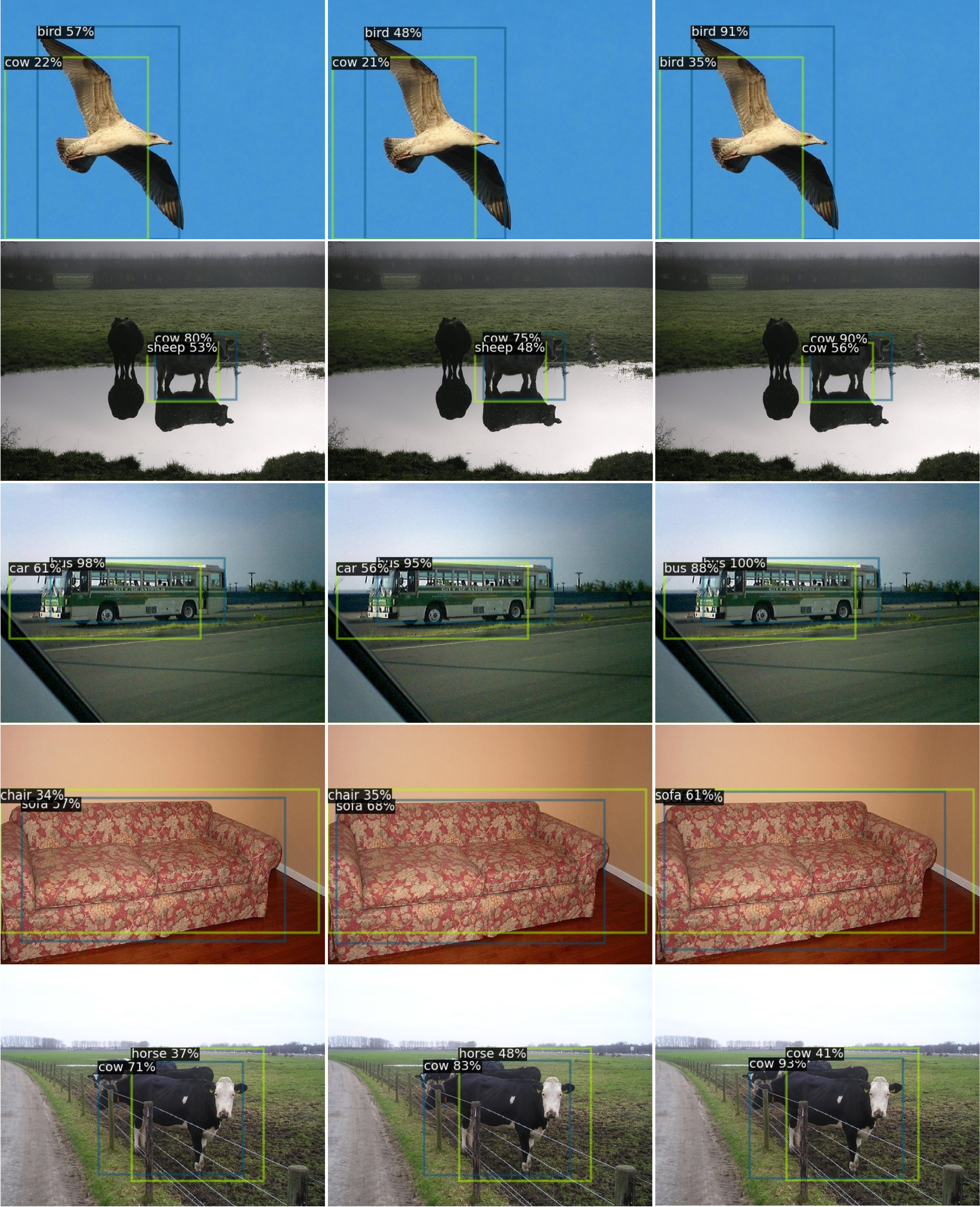}
\makebox[\subboxsize]{\hspace{1.2cm}DeFRCN \cite{Qiao2021DeFRCNDF}}
\makebox[\subboxsize]{Vanilla-VAE}
\makebox[\subboxsize]{\hspace{-1.2cm}Norm-VAE}
 \caption{ \textbf{Qualitative visualizations of the detected objects on PASCAL Novel Split 1}. ``Vanilla-VAE'' denotes the model trained with features generated from a vanilla VAE and ``Norm-VAE'' denotes the model trained with features generated from Norm-VAE. The blue box is the detector's prediction on the original image and the yellow box is the prediction on the augmented box. Our proposed Norm-VAE can generate features that enhance the model's robustness against crop-related variation.
}
\label{fig:vis_inacc}
\end{figure*}
In this section, we provide qualitative visualizations of the detected objects of the 1-shot model on PASCAL VOC Novel Split 1. As shown in Figure  \ref{fig:vis_inacc}, for each input image, the blue box is the original prediction result from the object detector. We then randomly create an augmented bounding box based on the ground-truth bounding box and input the augmented box to the classifier of the object detector. The prediction result on the augmented box is denoted as the yellow box. For the examples shown in the figure, the baseline DeFRCN model \cite{Qiao2021DeFRCNDF} and the model trained with features from a vanilla VAE predict the class labels correctly on the original input boxes while both fail on the augmented boxes. By contrast, the model trained with features from Norm-VAE can classify both the original box and the augmented box correctly. As can be seen, crop-related variation is crucial for object detection and our method can enhance the object detector's robustness against the variation successfully.

 \section{Number of Generated Samples}
 \label{sec:number_of_generated_samples}

In our main experiment, we generate 30 samples per class and use them together with the original few-shot samples to fine-tune the object detector. In this section, we investigate the impact of the number of the generated samples. Table \ref{tab:aug_nums} shows the AP50 on PASCAL VOC Novel Split 1 with different numbers of generated features under 1-shot, 2-shot and 3-shot settings. As the number of generated samples increases, the performance gradually improves and then plateaus and drops slightly (less than $0.5\%$ decrease in performance). 
\begin{table}[!ht] 
  \centering
\resizebox{0.6\textwidth}{!}{%
  \begin{tabular}{l|c|c|c|c|c|c|c|c}
    \toprule 
    \# Generated Features & 0 & 5 & 10 & 15 & 20 & 25 & 30 & 35 \\
    \midrule
    {1-shot} & {56.3} & {60.5} & {61.6} & {61.8} & {62.0} & {61.9} & \textbf{62.1} & {62.0}   \\
    {2-shot} & {60.3} & {62.0} & {63.7} & {63.6} & {63.6} & {64.1} & \textbf{64.9} & {64.5}  \\
    {3-shot} & {62.0} & {65.6} &  67.0 & 67.2 & 67.2 & \textbf{67.8} & \textbf{67.8} & 67.3
 \\
    \bottomrule
  \end{tabular}}\label{tab:lfw-ytf}  \\ \vspace{6pt}
  \caption{ \textbf{Impact of the number of the generated samples under PASCAL VOC Novel Split 1}. As the number of generated samples increases, the performance gradually improves and then saturates and drops slightly.
  }\label{tab:aug_nums}%
  \vspace{-10pt}
\end{table}

\section{Visualization of the Detection Results on PASCAL VOC dataset}
\label{sec:appendix_vis_2}

We show a few visualization results of DeFRCN \cite{Qiao2021DeFRCNDF} and our proposed method in Figure \ref{fig:detection_vis}. 
As can be seen from the figure, the model trained with additional features performs better than DeFRCN. For instance, in the third row, DeFRCN fails to recognize both the two instances of the ``\textit{bird}'' class while both Vanilla-VAE and Norm-VAE recognize them.  
It can be seen that with additional data from Norm-VAE, the FSOD model can recognize objects that are undetected with the model trained with just the original training data. The Norm-VAE model is generally more robust in recognizing objects. It works well even when the objects are cropped (2nd row) or small (two bottom rows).

\def\subFigSzab{\linewidth}
\def\subboxsize{0.3\textwidth}
\def\subfig{0.45\textwidth}
\def\W{0.8\textwidth}
\def\H{0.24\textwidth}
\begin{figure*}[!h]
 \centering
    \includegraphics[width=\W, ]{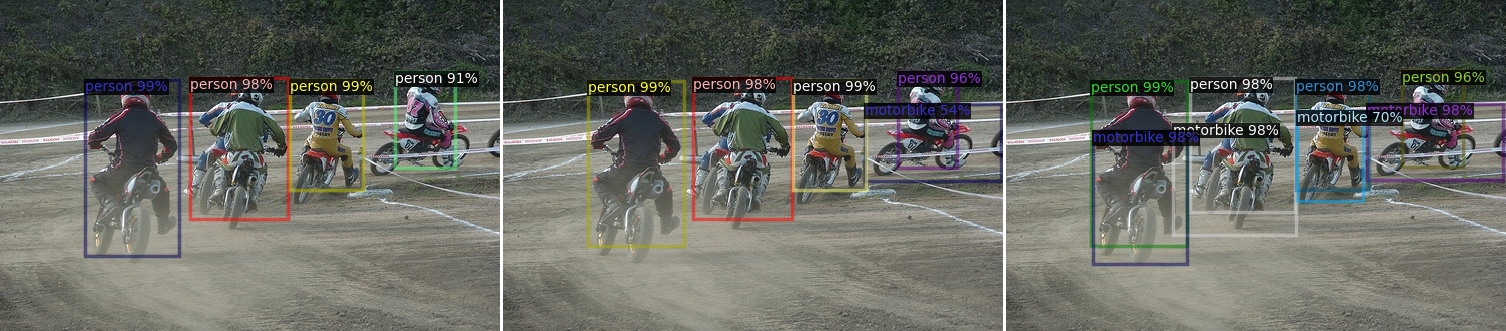}
    \includegraphics[width=\W]{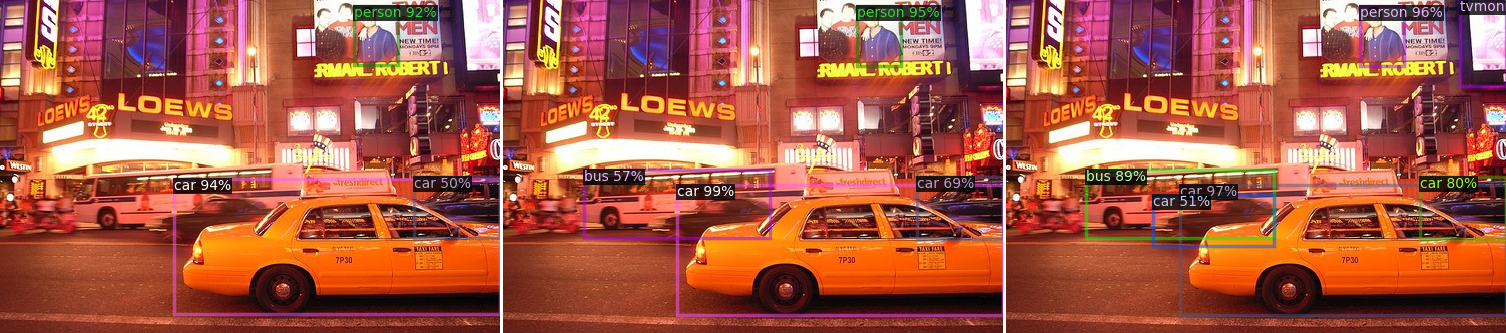}
     \includegraphics[width=\W]{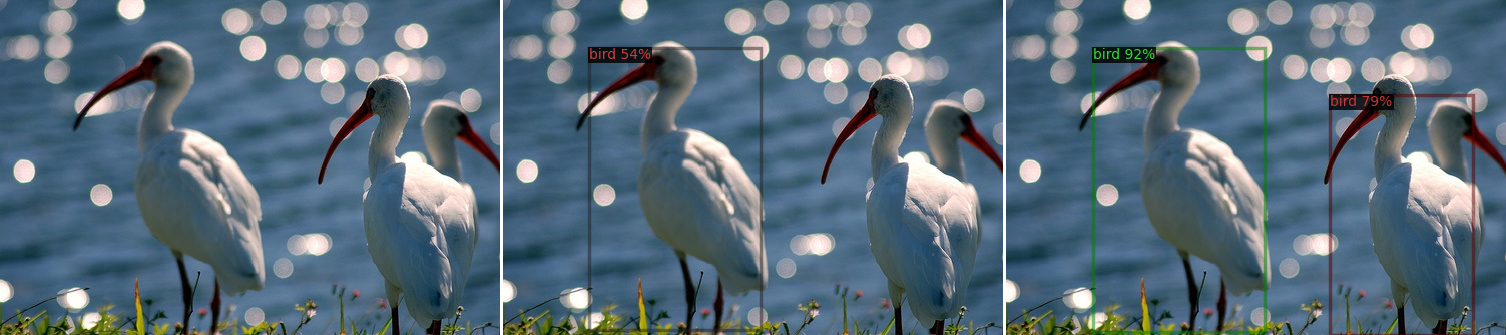}
    \includegraphics[width=\W]{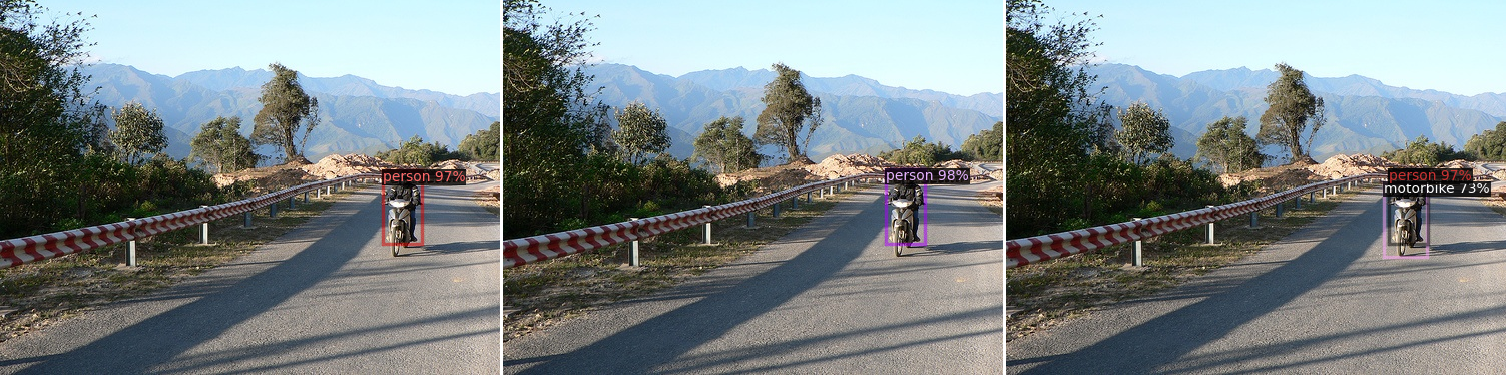}
    \includegraphics[width=\W, ]{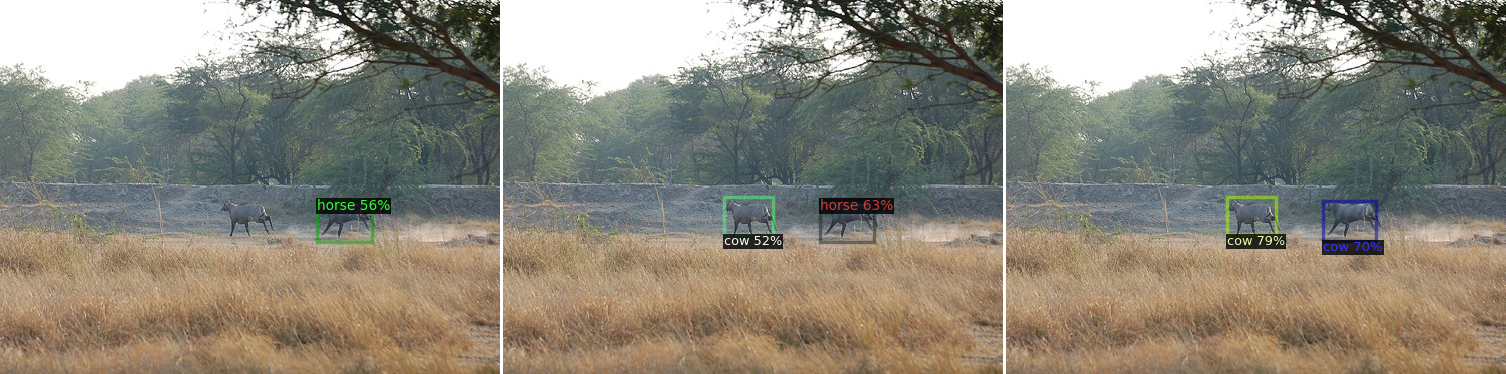}
    \makebox[\subboxsize]{\hspace{1.2cm}DeFRCN \cite{Qiao2021DeFRCNDF}}
    \makebox[\subboxsize]{Vanilla-VAE}
    \makebox[\subboxsize]{\hspace{-1.2cm}Norm-VAE}

\caption{\textbf{Visualization of the detection results on PASCAL VOC dataset.} 
The FSOD model trained with additional features performs better than DeFRCN. It works well even when the objects are partially cropped (2nd row) or small (two bottom rows). The detection score
threshold is 0.5. Please view in magnification for cases with small objects.}
\label{fig:detection_vis}
\end{figure*}

\section{Mapping Function Analyses}
\label{sec:mapping_function}
We use a simple pre-defined linear function $g(x) = w\times x + b$ to map from an IoU score $x$ to the new norm of a latent code. Here we only consider proposals with IoU scores ranging from 0.5 to 1. Proposals with lower IoU scores are noisy since they contain mostly background areas. 
With our VAE architecture and the training data, we observe that the norms of the original latent codes are ranged approximately from $\sqrt{512}$ to  $5\sqrt{512}$. 
We would like the rescaled norms to be in the same range and, at the same time, the latent code of an easy proposal has a small norm and the latent code of a hard proposal has a large norm.  Thus, we set the parameters of $g(x)$ such that $g(0.5) = 5\sqrt{512}$ and $g(1) = \sqrt{512}$.

We also conduct experiments with different ranges and the results are shown in Table \ref{tab:g_function}. Note that here $\sqrt{512}$ is a scaling constant that corresponds to the number of dimensions ($N = 512$) of the latent space. 
As can be seen from the table, we observe better performance when the IoU score inversely correlates with the latent norm. In this case, a proposal with a low IoU score (i.e., hard case)  has a higher latent norm and is placed further away from the origin. A possible reason is that features of hard instances often exhibit higher variance. Thus, it is more optimal to use latent codes with larger norms to represent them \cite{Meng2021MagFaceAU}.
\begin{table}[!h] 
  \centering
\resizebox{0.4\textwidth}{!}{%
  \begin{tabular}{c|ccc}
    \toprule
      & $g$(1) & $g$(0.5) & AP50 \\
    \midrule 
     \multirow{3}{*}{Inverse Correlation} & {1$\times\sqrt{512}$} & {2$\times\sqrt{512}$} & {61.6}  \\
     & {1$\times\sqrt{512}$} & {5$\times\sqrt{512}$} & \textbf{62.1}  \\
      & {1$\times\sqrt{512}$} & {10$\times\sqrt{512}$} & {61.8}  \\
    \midrule 
     \multirow{3}{*}{Correlation} & {2$\times\sqrt{512}$} & {1$\times\sqrt{512}$} & {60.6} \\
     & {5$\times\sqrt{512}$} & {1$\times\sqrt{512}$} & {61.3}  \\
      & {10$\times\sqrt{512}$} & {1$\times\sqrt{512}$} & {60.6}  \\
    \bottomrule
  \end{tabular}}
  \caption{\textbf{Performance with different configurations of the mapping function}. We conduct experiments using different coefficients for function $g(\cdot)$, which defines the value range of the new norm of the latent code.   
    }\label{tab:g_function}%
  \vspace{-10pt}
\end{table}

\section{Details on Generating Augmented Training Data}
\label{sec:augmented_data}
We extract the image features from image crops from the base classes and use them to train a feature generator to generate features for the novel classes. 
Specifically, we apply the RoI head feature extractor on the ground-truth bounding box $b_i$ from the base classes to get the RoI feature $f_i$. To enrich the diversity of the RoI feature, we randomly create $N$ additional augmented bounding boxes by randomly moving the starting point and the ending point of the original box, annotated as $\{b_{i}^1, b_{i}^2, ... b_{i}^N\}$.
These augmented bounding boxes overlap the ground-truth bounding box differently and have different IoU scores.
With a set of augmented bounding boxes $\{b_{i}^1, b_{i}^2, ... b_{i}^N\}$, we extract the corresponding RoI features $\{f_{i}^1, f_{i}^2, ... f_{i}^N\}$ and use them to train our VAE model.

\FloatBarrier
{\small
\bibliographystyle{ieee_fullname}
\bibliography{egbib,attributes_bib}
}